\documentclass[journal]{IEEEtran}

\usepackage{algorithm}
\usepackage{algorithmicx}
\usepackage{algpseudocode}

\usepackage{graphicx}
\usepackage{cite}
\usepackage{amsmath}
\usepackage{url}
\usepackage[latin1]{inputenc}
\usepackage{multirow}
\usepackage{pifont}
\usepackage{color}
\usepackage{alltt}
\usepackage[hidelinks]{hyperref}
\usepackage{enumerate}
\usepackage{siunitx}
\usepackage{epstopdf}
\usepackage[switch]{lineno}

\floatname{algorithm}{Workflow}

\begin{document}

\title{
\renewcommand\arraystretch{1.5}
Feature Analyses and Modelling of Lithium-ion Batteries Manufacturing based on Random Forest Classification}

\author{
	\vskip 1em
	Kailong Liu, Xiaosong Hu, Huiyu Zhou, Lei Tong, 
	W. Dhammika Widanage, James Marco	    

	\thanks{
	 \textit{(Corresponding author: Kailong Liu, Xiaosong Hu)}		
	
		K. Liu, D. Widanage and J. Marco are with the Warwick Manufacturing Group, University of Warwick, Coventry, CV4 7AL, and the Faraday Institution, Quad One, Harwell Science and Innovation Campus, Didcot, UK (e-mails: Kailong.Liu@warwick.ac.uk, kliu02@qub.ac.uk; Dhammika.Widanalage@warwick.ac.uk; James.Marco@warwick.ac.uk).
		
		X. Hu is with the Department of Automotive Engineering, Chongqing University, Chongqing 400044, China (e-mail: xiaosonghu@ieee.org).
		
	H. Zhou and L. Tong are with School of Informatics, University of Leicester, United Kingdom (e-mails: {hz143;lt228}@leicester.ac.uk).
  }
}

\maketitle

\begin{abstract}

Lithium-ion battery manufacturing is a highly complicated process with strongly coupled feature interdependencies, a feasible solution that can analyse feature variables within manufacturing chain and achieve reliable classification is thus urgently needed. This article proposes a random forest (RF)-based classification framework, through using the out of bag (OOB) predictions, Gini changes as well as predictive measure of association (PMOA), for effectively quantifying the importance and correlations of  battery manufacturing features and their effects on the classification of electrode properties. Battery manufacturing data containing three intermediate product features from the mixing stage and one product parameter from the coating stage are analysed by the designed RF framework to investigate their effects on both the battery electrode active material mass load and porosity. Illustrative results demonstrate that the proposed RF framework not only achieves the reliable classification of electrode properties but also leads to the effective quantification of both manufacturing feature importance and correlations. This is the first time to design a systematic RF framework for simultaneously quantifying battery production feature importance and correlations by three various quantitative indicators including the unbiased feature importance (FI), gain improvement FI and PMOA, paving a promising solution to reduce model dimension and conduct efficient sensitivity analysis of battery manufacturing.
\end{abstract}


{}

\definecolor{limegreen}{rgb}{0.2, 0.8, 0.2}
\definecolor{forestgreen}{rgb}{0.13, 0.55, 0.13}
\definecolor{greenhtml}{rgb}{0.0, 0.5, 0.0}

\section{Introduction}
\label{S1}
As a consequence of the manufacturing complexity that involves numerous individual process stages, a large number of variables and parameters are generated and coupled during battery manufacturing \cite{wood2019formation}. These process parameters will highly affect the properties of manufacturing intermediate products, which, in turn, further determine the final battery performance. Unfortunately, due to the complexity, the multiple inter-relations among key processes and control variables are still difficult to be understood. Currently the analysis of manufacturing variables to improve battery performance is still mainly dependent on the trial and error methods \cite{schmidt2020modeling}. Therefore, it is vital to develop  powerful data analysis solutions for better understanding and evaluating the variable importance, the process interactions within battery manufacturing chain.

With the rapid development of cloud computing and machine learning technologies, artificial intelligence and data-driven based strategies are becoming powerful tools in many industrial fields. For instance, a genetic algorithm and neural network based data-driven method was proposed in~\cite{ghahramani2020ai} for smart semiconductor manufacturing. In~\cite{saez2019context}, through considering the machine-interactions and operational context, a hybrid data-driven and physics-based framework was derived for modelling manufacturing equipment to improve anomaly detection and diagnosis. For battery applications, numerous data-driven models have been derived to estimate  operational states \cite{zhang2015novel,hu2019state,feng2020co,wang2020comprehensive}, predict service life \cite{li2019data,hu2020health,liu2019modified,liu2019gaussian}, diagnose faults \cite{hu2020advanced}, achieve effective charging \cite{ouyang2019optimal,zou2018model,liu2018charging} and energy managements \cite{shang2019compact,dai2019analytical}. However, all these researches mainly focus on the in-situ operation of battery performance without considering the microscopic properties of its production. As battery manufacturing also generates a large amount of data, it should also be a promising way by designing reliable data-driven solutions to analyse and improve  processes within it.
 
 In comparison with battery management research, fewer works have been done so far by applying machine-learning techniques in battery manufacturing domain. Among lots of corresponding themes (process monitoring \cite{knoche2016process}, adjustments \cite{schunemann2016smart} and analyses \cite{gunther2019classification}) of battery manufacturing, deriving suitable data-driven models to predict and analyse the intermediate products belongs to a significant research challenge. For instances, through analysing the initial failure mode and effect, Schnell et al. \cite{schnell2016quality} proposed a data-driven method for the internal decisions of battery manufacturing quality control without considering the link of each quality parameter. Then in \cite{schnell2019data}, a data-mining concept named the cross-industry standard process (CRISP) together with linear model, neural network, and regression approach are utilised to identify the process dependency and  predict the product qualities of battery manufacturing. According to the CRISP concept, Turetskyy et al. \cite{turetskyy2019toward} proposed a decision trees-based framework to conduct  manufacturing feature selection and regression models for predicting battery maximal capacity. In \cite{thiede2019data}, a multi-variate regression approach based on CRISP concept was also proposed to predict the final battery manufacturing properties and suggest the suitable quality gates. Based upon the defined capability indices, a hierarchical model was proposed in \cite{kornas2019multivariate} to determine performance indicators of production chain such as battery weight and capacity. Through using a statistical investigation of battery product fluctuations, Hoffmann et al. \cite{hoffmann2019capacity} investigated their effects on the manufactured cell capacities. In \cite{cunha2020artificial}, three common data-driven models including support vector machine, decision tree and neural network are utilised to classify the electrode properties. Then the parameter dependencies are analysed through the 2D graphs from model and experiment data.  For the aforementioned applications,  reasonable data-driven analyses of battery manufacturing have been obtained, and several limitations still exist as: 1) researches mainly focus on simply using the existed common methodologies to predict battery product properties, lacking of in-depth investigations on the characteristics of adopted machine learning techniques to further enhance their performance and generalization in battery manufacturing domain. 2) many works mainly emphasize the accuracy of developed model, ignoring systematically analysing its interpretability for battery manufacturing data. For the battery production chain that presents various feature variables, apart from obtaining the predicted output of utilised model, manufacturers are also very interested in the underlying correlations among different variables and which features are more crucial for determining the predicted results. Such information can effectively help battery manufacturers optimise their battery products.
 
Based upon the above discussions, it becomes significantly meaningful to design the interpretable model for effectively predicting battery manufacturing outputs with reliable intermediate feature analyses being taken into account. To achieve this, a novel data-driven framework based on the improved random forest (RF) classification technique is designed in this study to simultaneously classify battery electrode properties and determine the levels of both feature importance as well as correlations. Specifically, some key contributions are made as follows: 1) according to a well labeled battery electrode manufacturing dataset with 5 classes, effective RF model structure with the bagging  and OOB prediction solutions is designed, bringing the benefits to achieve unbiased classification of battery electrode properties and highly restrain the overfitting phenomenon. 2) through randomly permuting feature observations within OOB and calculating the Gini changes, two different types of feature importance (FI) including both unbiased FI as well as gain improvement FI can be derived to directly quantify the importance levels of selected mixing and coating features. 3) a powerful noise immunity solution named PMOA is designed from the surrogate decision split, which is able to effectively quantify the strength of correlations between all pairs of manufacturing feature variables. 4) the developed RF-based approach is analysed in-depth to evaluate the effects of four key variables from the mixing and coating stages on the classifications of two battery product properties - electrode mass load and porosity. Obviously, through using the proposed RF-based framework, the importance and correlations of all manufacturing feature variables can be well quantified and analysed. This is the first known application of designing a systematic RF-based framework to not only classify the electrode properties but also quantify the importance and correlations of involved mixing and coating features with three different evaluation criteria. Due to the data-driven nature, this framework can be conveniently extended to other processes of battery manufacturing chain after collecting the available data, paving a promising way for the  reliable sensitivity analysis of intermediate features and the improvement of model dimension as well as battery manufacturing process.

The remainder of this article is organised as follows. Section~\ref{S2} specifies the battery manufacturing chain and several key process steps. Then the fundamentals behind the RF classification technology, feature importance/correlation determination, classification model structure and framework, as well as performance metrics are described in Section~\ref{S3}. Section~\ref{S4} details  our classification results with the in-depth discussions of feature correlations and importance. Finally, Section~\ref{S5} summaries the conclusion of present work.

\section{Battery manufacturing fundamentals}
\label{S2}

\begin{figure*}[ht]
		\centering
		\includegraphics[angle=0,
		width=0.6\textwidth]{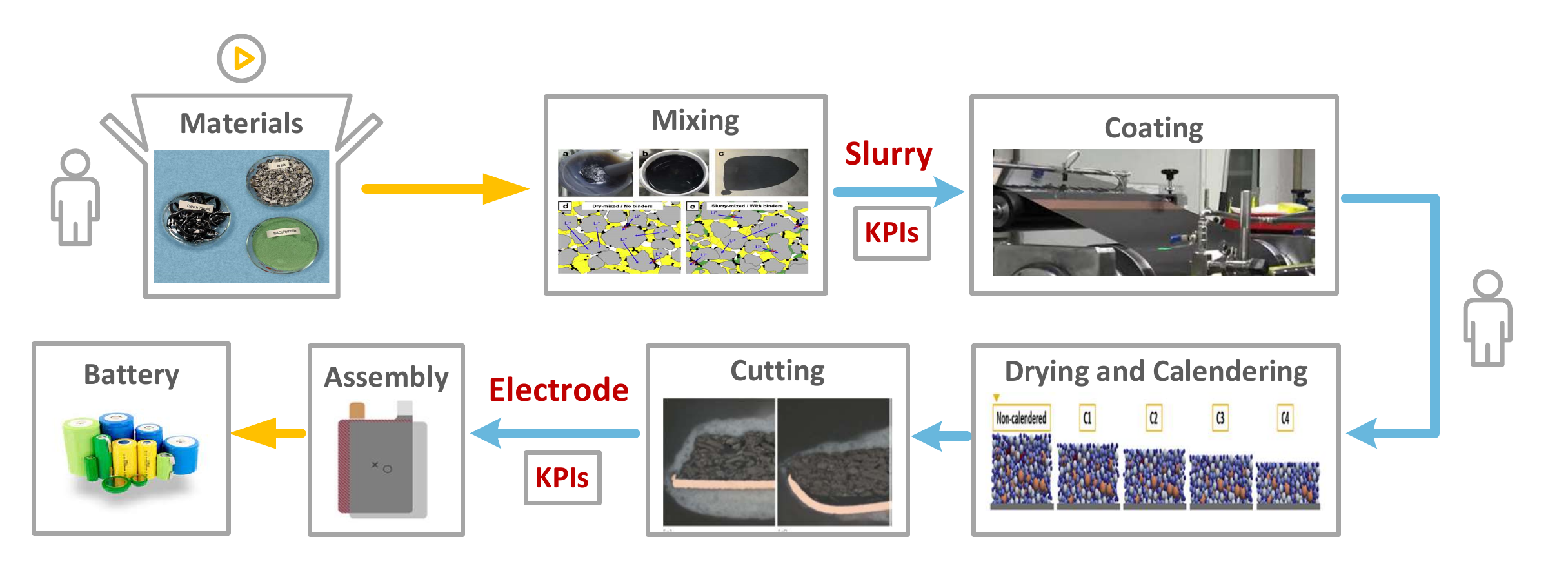}   
		\caption{Key processes within battery production chain especially for electrode manufacturing. }
		\label{fig:manufacturingprocess} 
	\end{figure*}

Li-ion battery manufacturing is a long and highly-complicated process chain, which mainly consists of electrodes manufacturing, cell assembly, formation and ageing. Fig.~\ref{fig:manufacturingprocess} systematically illustrates several key intermediate processes within the battery production chain especially for electrode manufacturing. In general, after preparing active materials, the slurry could be made within a soft blender through a mixing stage. Then the slurry is coated on the surface of copper or aluminium foils by a comma-gap coater with several built-in ovens to dry the coating products. Afterwards, the anode and cathode electrodes are obtained through calendering and cutting the dried coating products. Then all components such as electrodes and electrolyte are assembled to produce the basic battery cell. Due to the highly complicated operations within battery production chain, engineers can control the electrode mass load and porosity more conveniently and easily with the discrete data and class form in real battery manufacturing \cite{kwade2018current}. An effective classification approach could thus benefit battery manufacturer in such a case.

In this context, to design a reliable RF-based classification framework for analysing the feature importance and correlations of battery electrode manufacturing, some key IPFs and PPs from mixing and coating stages are studied in this article. Besides, their effects on the classification performance of battery electrode characteristics are also investigated. Without the loss of generality, the whole raw dataset \cite{cunha2020artificial} from Franco Laboratoire de Reactivite et Chimie des Solides (LRCS)  is explored in this study, which leads the total number of feature variables here is four. Specifically, these interested battery manufacturing features including three slurry IPFs (active material mass content (AMMC), solid to liquid ratio (StoLR) and viscosity) as well as one coating PP (comma-gap (CG)). The StoLR reflects the mass ratio among slurry solids and slurry mass. Viscosity affects the shear rate of coating step. CG represents the gap between comma and coating rolls. For the battery electrode characteristics, two key variables including the electrode mass load with unit $mg/cm^2$ and porosity after drying with unit \% are utilised to reflect the electrode product properties. Detailed information regarding the experiments and data explanations can be found in \cite{cunha2020artificial}, which is not repeated here due to space limitations. For this raw dataset with 656 samples, eight same samples of slurry IPFs and coating CG are used to generate one related electrode mass load and porosity. Therefore, 82 observations are generated by averaging the related eight samples. To fully investigate the effectiveness of RF classification, both electrode mass load and porosity are classified into multi-classes with five labels (very low, low, medium, high and very high), respectively. The detailed class label setting rules are illustrated in Table~\ref{table:classrules}.

\begin{table}[h]
		\caption{Class label setting rules of battery electrode mass load and porosity} 
		\centering
		\begin{tabular}{p{1.4cm}<\centering  p{3cm}<\centering  p{2.2cm}<\centering   
				} 
			\hline\hline   
			\rule{-3pt}{1.0\normalbaselineskip}
			Class labels      & Mass load [mg/$cm^2$] & Porosity [\%] \\ 
			\hline		    
			very low	 & $\le 15$	 & $\le 47.5$ \\
			low  & $15< ML\le 25$  & $47.5<Po \le 50$  \\
			medium	 & $25<ML\le 35$ & $50<Po\le 52.5$\\
			high 	 & 	$35<ML\le 45$  & $52.5<Po\le 55$	\\ 
						very high	 & $>45$ & $>55$ \\
			\hline\hline 	
		\end{tabular}
		\label{table:classrules}
	\end{table}

	\begin{figure}[h]
		\centering
		\includegraphics[angle=0,
		width=0.42\textwidth]{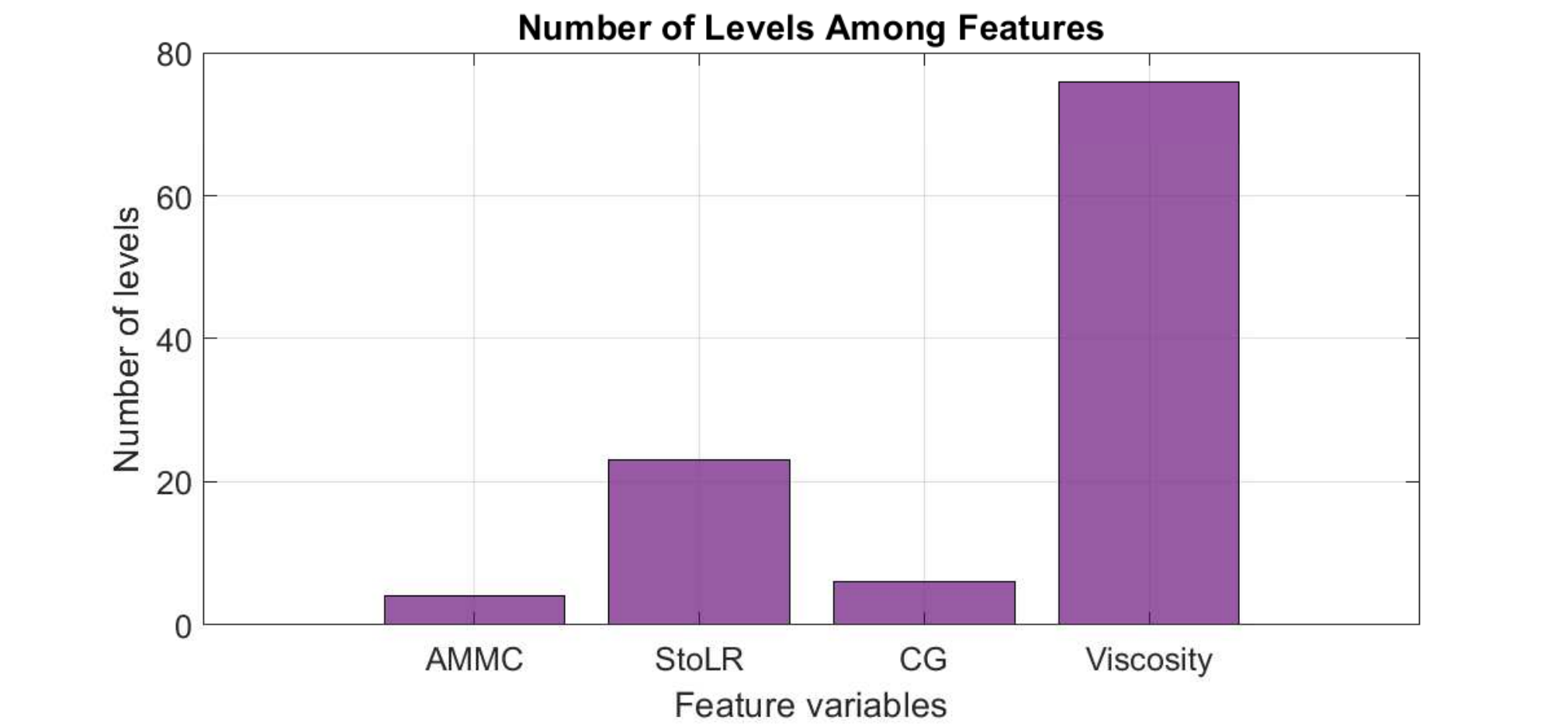}  
		\caption{The number of levels among all interested features.}
		\label{fig:Nofeatures} 
	\end{figure}

	Fig.~\ref{fig:Nofeatures} details the number of levels of each features. Obviously, viscosity belongs to a continuous variable with 76 number levels, which is significantly more than other three features (here the number levels of AMMC, StoLR and CG are 4, 23 and 6, respectively). Based upon these feature data with large different number levels and preset class labels, the RF-based classification framework is then designed to analyse the  importance and correlations of these features in this study.
	
\section{Methodology}
\label{S3}
This section first describes the fundamental of RF. Then the process to conduct feature analyses is elaborated, followed by the description of RF-based framework to classify battery electrode mass load and porosity. Additionally, the performance metrics to evaluate classification results are also given.

\subsection{Random Forest}
	\begin{figure}[h]
		\centering
		\includegraphics[angle=0,
		width=0.32\textwidth]{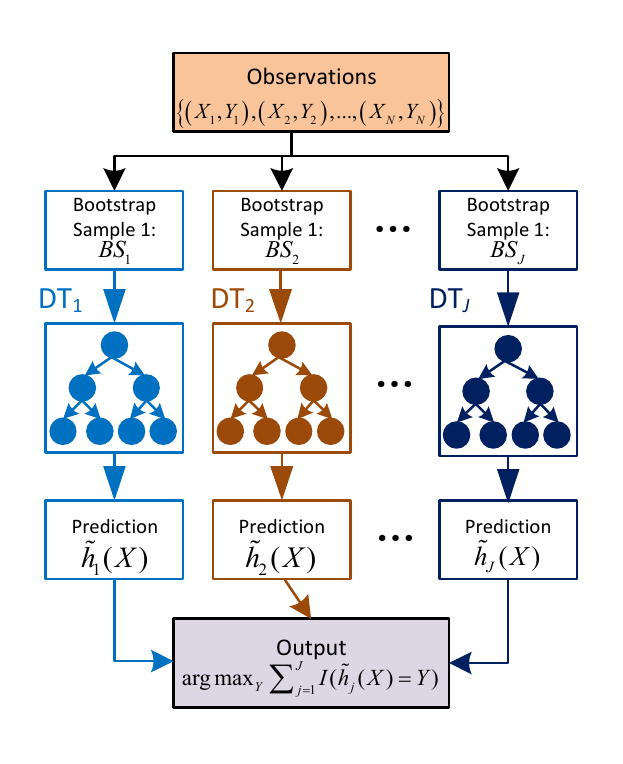}  
		\caption{Structure of RF classification model.}
		\label{fig:RFstructure} 
	\end{figure}
 
Derived from ensemble learning theory, RF combines multiple individual decision trees (DTs)~\cite{cutler2012random}. Due to the simplification and nonparametric behaviours, classification and regression tree (CART) is generally utilised as a DT within RF\cite{liaw2002classification}. Each DT relies on a random bootstrap dataset. The structure of RF classification model is shown in Fig.~\ref{fig:RFstructure}~\cite{li2018random}. For the classification issue, supposing training data $TD=\{(X_1,Y_1),(X_2,Y_2),...,(X_N,Y_N)\}$ contain $N$ observations, $X_i$ stands for the input vector owing $M$ features as $X_i=(x_{i1},x_{i2},...,x_{iM})$, $Y_i$ is the output scalar, the process of establishing a RF classification model is detailed in Workflow~\ref{alg:RF}.

\begin{algorithm}[!htb]
\caption{Detailed process to establish RF-based classification model}             
\label{alg:RF}                        
\begin{algorithmic}[1]
\Procedure{RF Training}{}

\State For $j=1$ to $J$: ($J$ is the number of DTs)

\State Formulate a bootstrap sample $BS_j$ with $N$ observations from $TD$;

\State Fit a tree $DT_j$ based on its $BS_j$:

a. Start splitting a node with all observations of $BS_j$.

b. Recursively repeat the following processes on each unsplit node: 

  \quad i. Randomly choose $m$ features $(m < M)$ from $M$ candidates: 
$ m \leftarrow M$

  \quad ii. Discover the split solution with the best impurity among all possible splits of $m$ features  from Process i. 

  \quad iii. Split this node into two sub-nodes based on the obtained split solution from Process ii.
 
 \State Obtain the well-trained RF through ensembling all base DT learners $h_j(.)$. 
  
 \EndProcedure 
  
  \Statex

\Procedure{RF Classification}{}

\State For a new observation $X_{new}$, the output $RF(X_{new})$ of RF is predicted by:

\quad $RF({X_{new}}) = \arg {\max _Y}\sum\nolimits_{j = 1}^J {I({{\tilde h}_j}({X_{new}}) = Y)}$

where ${{{\tilde h}_j}({X_{new}})}$ is the $j$th DT's prediction result with  $X_{new}$ as inputs. $I(.)$ is a zero-one judgement with  ${I({{\tilde h}_j}({X_{new}}) = Y)}=1$. $\arg {\max _Y}$ outputs the class with the maximum counting number from all DTs.
\EndProcedure
\end{algorithmic}
\end{algorithm}

The main purpose of the RF training stage is to construct numerous de-correlated DTs. To decrease the variance associated with classification, an overlap sampling solution named 'bagging' is adopted in the RF \cite{li2018random}. Specifically, it extracts observations with replacement to generate the independent bootstrap sample from training dataset. Then each decision tree can be trained from different bootstrap samples, leading to an increased tree diversity. Besides, to further restrain the correlations among various DTs, the best split of each node is obtained through randomly selecting $m$ subset features instead of all $M$ features. As a result, DTs within RF can be grown without pruning, leading to a relatively small computational burden. Moreover, through using different bootstrap samples and node features, the noise immunity of RF can be improved with the help of averaging various de-correlated DTs. 

Additionally, for each DT within a RF, due to the bagging solution, some training data would be repeatedly utilised as the bootstrap sample, resulting in some other observations not being selected to fit this DT. These observations are  named as out of bag (OOB) samples. In general, nearly one-third TD constitutes OOB samples and would not be utilised in RF training process. Therefore, at each time when a DT has been trained, the OOB samples can be used to evaluate the classification performance of this DT. In this way, RF is able to achieve unbiased estimations without using external data subset. For the classification of battery product properties, OOB predictions with the related generalization error $E_{OOB}$ of RF can be obtained by  the workflow~\ref{alg:OOBpredictions} below.

\begin{algorithm}[!htb]
\caption{OOB predictions and the generalization error}             
\label{alg:OOBpredictions}                        
\begin{algorithmic}[1]
\Procedure{OOB predictions}{}

\State For $i=1$ to $N$:

i. Suppose $A_i = \left\{ {j:({X_i},{Y_i}) \notin B{S_j}} \right\}$, and $J_i$ is the cardinality of $A_i$.

ii. Obtain the OOB prediction at $X_i$ by:

\quad ${{\tilde f}_{OOB}}({X_i}) = \arg {\max _Y}\sum\nolimits_{j \in {A_i}} {I\left( {{{\tilde h}_j}({X_i}) = Y} \right)}$

where ${{{\tilde h}_j}({X_i})}$ is the prediction result by using $X_i$ as inputs to the $j$th DT.

\State Calculate the generalization error $E_{OOB}$ by:

\quad ${E_{OOB}} = \frac{1}{N}\sum\nolimits_{i = 1}^N {I({Y_i} \ne {{\tilde f}_{OOB}}({X_i}))} $

\EndProcedure
\end{algorithmic}
\end{algorithm}

It should be noticed that the final $E_{OOB}$ is calculated through the error rate of OOB predictions rather than averaging each DT's OOB error. In light of this, a class-wise error is obtained for each class, while a confusion matrix for the classification of battery manufacturing could be also generated.

\subsection{Feature Importance and correlation}

To effectively quantify the importance of both mixing features and coating parameter of battery production, the unbiased FI that obtained by OOB prediction is first utilised. Detailed process to obtain the unbiased FI is shown in Workflow~\ref{alg:OOBimportance}.

\begin{algorithm}[!htb]
\caption{Unbiased FI based on OOB predictions}             
\label{alg:OOBimportance}                        
\begin{algorithmic}[1]
\Procedure{To estimate the unbiased importance of features $x_k$}{$k=1$ to $M$}

    \State (Obtain $\tilde Y_{i,j}$) For $i=1$ to $N$:

i. Suppose $BS_j$ is the $j$th bootstrap sample, $A_i = \left\{ {j:({X_i},{Y_i}) \notin B{S_j}} \right\}$, and $J_i$ is the cardinality of $A_i$.

ii. Obtain ${{\tilde Y}_{i,j}} = {{\tilde h}_j}({X_i})$ for all $j \in A_i$.

\State (Obtain $\tilde Y{'_{i,j}}$) For $j=1$ to $J$:

i. Suppose ${B_j} = \left\{ {i:({X_i},{Y_i}) \notin B{S_j}} \right\}$ 

ii. Randomly permute $x_k$ from data samples $\left\{ {{X_i}:i \in {B_j}} \right\}$ to generate ${C_j} = \left\{ {X{'_i}:i \in {B_j}} \right\}$.

iii. Obtain $\tilde Y{'_{i,j}} = {{\tilde h}_j}(X{'_i})$ for all $i \in {B_j}$.

\State For $i=1$ to $N$:

Calculate the local FI $LFM_i(x_k)$ of $x_k$ as:

\quad $\begin{array}{*{20}{c}}
{LF{M_i(x_k)} = \frac{1}{{{J_i}}}\sum\nolimits_{j \in {A_i}} {I\left( {{Y_i} \ne \tilde Y{'_{i,j}}} \right)} }\\
{ - \frac{1}{{{J_i}}}\sum\nolimits_{j \in {A_i}} {I\left( {{Y_i} \ne {{\tilde Y}_{i,j}}} \right)} }
\end{array}$

\State Obtain the overall unbiased importance ($OFM_{xk}$) of feature $x_k$ as:

\quad $OF{M_{xk}} = \frac{1}{N}\sum\nolimits_{i = 1}^{i = N} {LF{M_i(x_k)}} $

\EndProcedure
\end{algorithmic}
\end{algorithm}

In this workflow, $LFM_i(x_k)$ is calculated by averaging over observations with size $J_i$ from the same class, while $OFM_{xk}$ is obtained through averaging over all observations with size $N$. Therefore the unbiased importance of feature $x_i$ could reflect how much the classification error varies when the values of $x_i$ are randomly permuted in the OOB prediction tests. 

Apart from the unbiased FI, another effective solution to evaluate the importance of features is through summing the gain improvements of Gini impurity changes caused by the splits on each feature. For the classification, Gini impurity is utilised to measure how well a potential split is in a specific node of DT \cite{nembrini2018revival}. Detailed process to obtain the gain improvement FI is illustrated in Workflow~\ref{alg:Giniimportance}. Obviously, ${I_G}({x_k})$ could reflect the gain improvement from the splits of feature $x_k$. Larger value of ${I_G}({x_k})$ indicates that this $x_k$ brings higher impurity improvement for the target classification.

\begin{algorithm}[!htb]
\caption{Gain improvement FI based on Gini changes}             
\label{alg:Giniimportance}                        
\begin{algorithmic}[1]
\Procedure{To estimate the gain improvement FI of features $x_k$}{$k=1$ to $M$}

    \State (Obtain $\Delta Gini(\tau,x_k)$) For $j=1$ to $J$:

i. For a node $\tau$ of $DT_j$, calculate its Gini impurity $Gini(\tau)$ by:

\quad $Gini(\tau) = 1 - \sum\nolimits_{d = 1}^D {p_k^2}$

where $D$ is the number of classes, $p_k={n_k}/n$ is the fraction of $n_k$ samples out of total $n$ samples.

ii. Calculate all Gini impurities $Gini(\tau,x_i)$ under the case of selected feature $x_i$ by:

\quad $Gini(\tau ,X) = \frac{{\left| {{\tau _l}} \right|}}{{\left| \tau  \right|}}Gini({\tau _l}) + \frac{{\left| {{\tau _r}} \right|}}{{\left| \tau  \right|}}Gini({\tau _r})$.

where $\tau _l$ and $\tau _r$ are the left child and right child of the current node $\tau$, respectively; ${\left| {{\tau}} \right|}$, ${\left| {{\tau _l}} \right|}$ and ${\left| {{\tau _r}} \right|}$ represent the number of records in $\tau$, $\tau _l$ and $\tau _r$, respectively.

iii. Calculate the Gini decrease $\Delta Gini(\tau ,X)$ of all selected $X$ by:

\quad $\Delta Gini(\tau ,X)=Gini(\tau )-Gini(\tau ,X)$

iiii. Compare $\Delta Gini(\tau ,X)$ to obtain the optimal split feature $x_k$ at this specific node $\tau$. Record its Gini decrease $\Delta Gini(\tau,X_k)$.

\State (Obtain ${I_G}({x_k})$) For $j=1$ to $J$:

i. Accumulate the recorded $\Delta Gini(\tau,x_k)$ for all used nodes (ANs) in all trees (ATs) by:

\quad $S\Delta Gini({x_k}) = \sum\limits_{ATs} {\sum\limits_{ANs} {\Delta Gini(\tau ,{x_k})} }$

where $S\Delta Gini({x_k})$ is the summed gain improvement based on $x_k$'s Gini changes.

ii. Calculate the overall gain improvement ${I_G}({x_k})$ of feature $x_k$ as:

\quad ${I_G}({x_k}) = \frac{1}{{{N_{xk}}}}S\Delta Gini({x_k})$

where ${{N_{xk}}}$ is the cardinality of $S\Delta Gini({x_k})$.

\EndProcedure
\end{algorithmic}
\end{algorithm}

On the other hand, evaluating the correlations among various electrode features is also crucial for better understanding battery manufacturing. To achieve this, an effective solution named the predictive measure of association (PMOA) is designed in this study. In theory, the value of PMOA could reflect the similarities between different decision rules to split observations. The basic idea of obtaining PMOA is to compare all potential splits with the optimal one that is founded by training DT. Then the best surrogate decision split would generate the maximum PMOA value, which could reflect the correlations between pairs of these two features. Supposing $x_e$ and $x_g$ are two interested feature variables ($e \ne g$), the detailed equation to calculate PMOA between the optimal split $x_e<u$ and surrogate split $x_g<v$ is expressed as follows:
\begin{align}
PMO{A_{e,g}} = \frac{{\min (Pl,Pr) - 1 + P{l_e}{l_g} + {{Pr }_e}{r_g}}}{{\min (Pl,Pr )}}
    \label{eq:PMOA}
\end{align}
where the subscripts $l$ and $r$ represent the left and right children of node, respectively; $Pl$ stands for the observation proportion of  $x_e<u$; $Pr$ is the observation proportion of $xe\ge u$; $P{l_e}{l_g}$ means the observation proportion of $x_e<u$ and $x_g<v$, while $P{r_e}{r_g}$ represents the observation proportion of $x_e\ge u$ and $x_g\ge v$. 
For the PMOA, the observations with several missing values of $x_e$ and $x_g$ would not affect the proportion results. $x_g<v$ could be selected as a worthwhile surrogate split for $x_e<u$ when $PMO{A_{e,g}}>0$. Besides, the range of PMOA should be within $( -\infty ,1]$, larger PMOA indicates more highly correlated pairs of feature variables.

\subsection{Classification model structure and framework}

\begin{figure*}[h!]
		\centering
		\includegraphics[angle=0,
		width=0.7\textwidth]{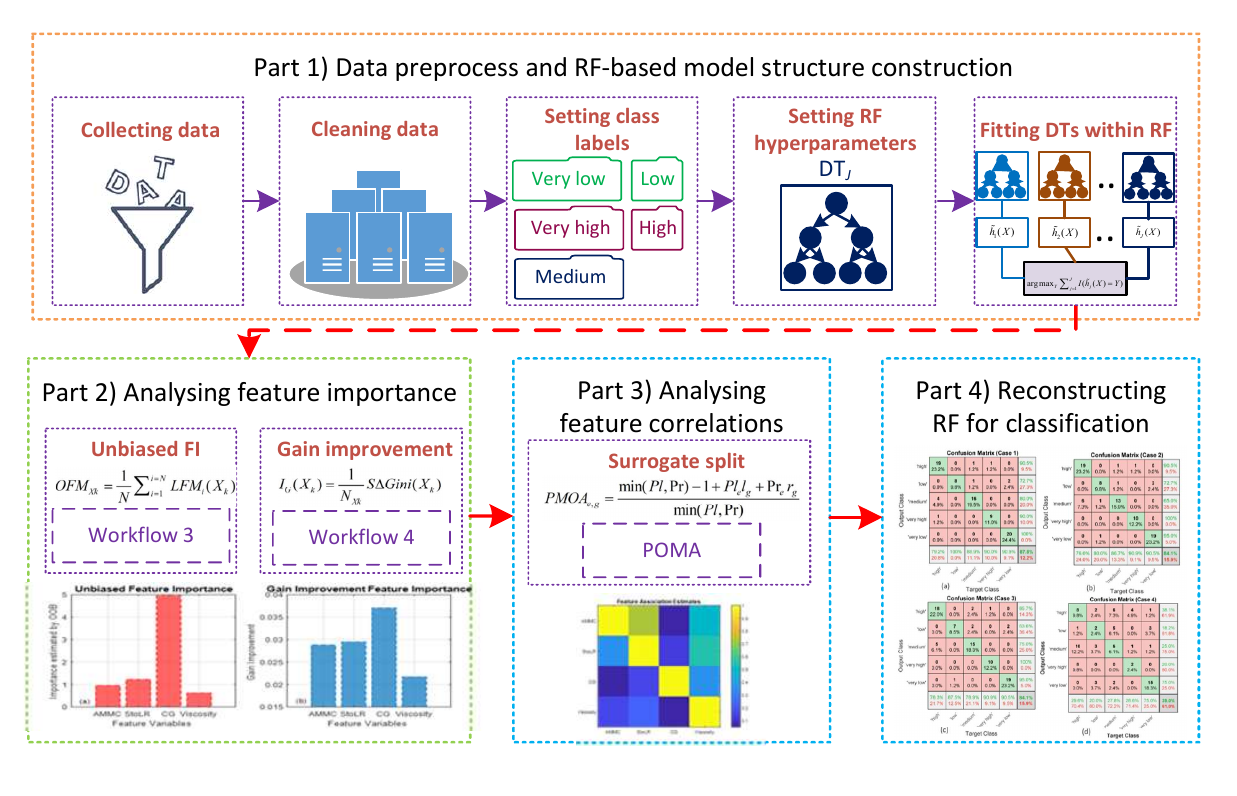}   
		\caption{Total framework to design RF-based model for classifying and analysing features importance as well as correlations.}
		\label{fig:RFclassificationworkflow} 
	\end{figure*}

For battery manufacturing process, mixing and coating are two key processes to affect electrode properties, further determining the performance of final manufactured battery \cite{kwade2018current}. To effectively quantify the FI and correlations among all interested variables, a RF classification model with the structure in Fig.~\ref{fig:RFmodel} is utilised. Specifically, the IPFs of mixing including AMMC, StoLR and viscosity of slurry, as well as one PP of coating named CG are utilised as the inputs, while the output of RF is the labelled classes of electrode mass load or porosity. Fig.~\ref{fig:RFclassificationworkflow} illustrates the total framework to design a RF model for classifying and analysing the FI as well as feature correlations under the specific inputs-output pairs of battery manufacturing. This framework consists of four main parts and is detailed as follows:

	\begin{figure}[h]
		\centering
		\includegraphics[angle=0,
		width=0.42\textwidth]{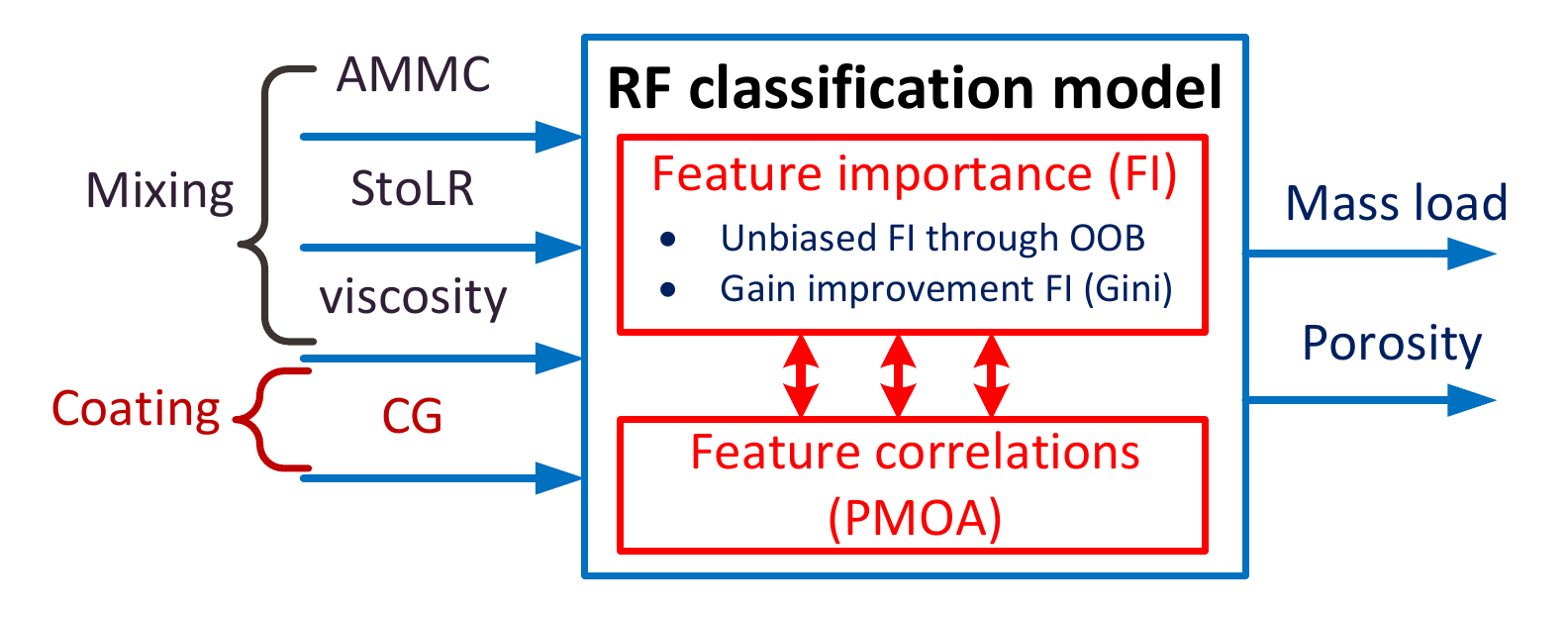}  
		\caption{RF-based classification model structure.}
		\label{fig:RFmodel} 
	\end{figure}

\textbf{Part 1) Data preprocess and RF-based model structure construction:} after collecting interested battery manufacturing data, the obvious outliers of original data are firstly removed and the suitable class labels of outputs are set. In our study, both the battery electrode mass load and porosity are classified with five class labels. Then the preprocessed inputs-output observations are utilised to train all DTs within RF through the steps in Workflow 1. As RF model is a powerful but easy-to-use machine learning method with only two hyper-parameters (the number of DTs ($J$) and the amount of features in each split ($m$)) to tune, some key points should be considered in this stage. First, for $J$, in theory, higher the number of DTs a larger accuracy and generalization ability is obtained. However, too many DTs would highly increase the computational effort of RF. Second, $m$ would affect the performance of each DT and the correlations among any DTs within RF. Large $m$ benefits the strength of each DT but also makes all DTs become more correlated. In our study, these hyper-parameters are tuned by an effective method named randomized search~\cite{bergstra2012random}.

\textbf{Part 2) Analysing feature importance:} in this part, to quantify all interested FI and analyse their effects on the classification performance of electrode mass load and porosity, two effective quantitative indicators including the unbiased FI and the gain improvement FI are utilised. Specifically, the unbiased FI is calculated by permuting OOB observations with the detailed process in Workflow 3, while the gain improvement FI is obtained by summing Gini changes caused by splits on each feature (Workflow 4). 

\textbf{Part 3) Analysing feature correlations:} after quantifying the importance of mixing and coating features, the PMOA values of each feature pair are calculated by equation (\ref{eq:PMOA}) and plotted as a $M \times M$ heat map. Then the correlations between each two features can be analysed by these PMOAs. In theory, larger PMOAs indicate there exists more highly correlations between feature pairs. In the heat map, the PMOAs of two features would be different, depending on which feature firstly generates the optimal spit within DTs.

\textbf{Part 4) Reconstructing RF for classification:} after comparing the FI and analysing feature correlations, the most important features that affect classification results are selected. Then the RF can be reconstructed with reduced feature set for new classifications.

Following this framework, an effective RF model-based framework can be formulated to not only analyse the importance and correlations of mixing and coating features, but also well classify the manufactured battery electrode mass load and porosity into suitable categories. Besides, after collecting more PPs, IPFs and product properties of a battery manufacturing chain, this framework can be further extended to analyse data correlations, discover most important features and simplify model structure with reduced variable set.

\subsection{Performance metrics}

In this subsection, to compare and quantify the classification performance of the designed RFs, several performance metrics including the confusion matrix, macro-precision, macro-recall as well as macro F1-score are applied in this study. 

In classification applications, let positive corresponds to the interested class while negative corresponds to other classes, four basic measures including true positives (TP), false positives (FP), true negatives (TN) and false negatives (FN) can be formulated for each class. For an interested class $c_h$ (here $h=1:5$), the precision rate ($P rate$) can be used to quantify the correct classification results of this class as:
\begin{align}
P rate(c_h)= {{TP} \mathord{\left/
 {\vphantom {{TP} {(TP + FP)}}} \right.
 \kern-\nulldelimiterspace} {(TP + FP)}}.
    \label{eq:Prate}
\end{align}

Recall rate ($R rate$) could quantify the rate of all fraud conditions of this class as:
\begin{align}
R rate(c_h) = {{TP} \mathord{\left/
 {\vphantom {{TP} {(TP + FN)}}} \right.
 \kern-\nulldelimiterspace} {(TP + FN)}}.
    \label{eq:Rrate}
\end{align}

F-measure ($F measure$) reflects the harmonic mean of precision and recall of this class as:
\begin{align}
Fmeasure({c_h}) = \frac{{2 \times Prate({c_h}) \times Rrate({c_h})}}{{Prate({c_h}) + Rrate({c_h})}}.
    \label{eq:Fmeasure}
\end{align}

The overall correct classification rate ($OCCrate$) to reflect the proportion of correctly classified observations out of all the observations is calculated by: 
\begin{align}
OCCrate = \frac{{T{P_{all}} + T{N_{all}}}}{N},
    \label{eq:microF1}
\end{align}
where $T{P_{all}} + TN_{all}$ represents all outputs that have been correctly classified, $N$ is the total number of observations.

Based upon the above mentioned metrics, a $(M + 1) \times (M + 1)$ confusion matrix (CM) of multi-class issue could be formulated. Each row within CM reflects the predicted output class while each column stands for the actual target class. The elements on the primary diagonal are the correct results while other elements reflect the incorrect classification cases. The ($M+1$)th column and row represent the $Prate(c_h)$ and $Rrate(c_h)$ of each class, respectively. The last element in the bottom right corner represents the $OCCrate$.

Supposing each class has a $Prate(c_h)$, $Rrate(c_h)$ and $Fmeasure(c_h)$, then the macro-precision ($macroP$), macro-recall ($macroR$) and macro F1-score ($macroF1$) can be calculated to evaluate the overall classification performance of our battery manufacturing multi-class issue as:
\begin{align}
\left\{ {\begin{array}{*{20}{c}}
{{macroP} = {{\sum\nolimits_{h = 1}^5 {Prate({c_h})} } \mathord{\left/
 {\vphantom {{\sum\nolimits_{h = 1}^5 {Prate({c_h})} } 5}} \right.
 \kern-\nulldelimiterspace} 5}}\\
{{macroR} = {{\sum\nolimits_{h = 1}^5 {Rrate({c_h})} } \mathord{\left/
 {\vphantom {{\sum\nolimits_{h = 1}^5 {Rrate({c_h})} } 5}} \right.
 \kern-\nulldelimiterspace} 5}}\\
{macroF1 = {{\sum\nolimits_{h = 1}^5 {Fmeasure({c_h})} } \mathord{\left/
 {\vphantom {{\sum\nolimits_{h = 1}^5 {Fmeasure({c_h})} } 5}} \right.
 \kern-\nulldelimiterspace} 5}}.
\end{array}} \right.
    \label{eq:macroPRF1}
\end{align}

\section{Results and discussions}
\label{S4}
To well quantify feature importance, feature correlations and their effects on the classification of electrode properties, the designed RF-based framework is utilised to classify both battery electrode mass load and porosity in this section. 

\subsection{RF classification model for battery mass load}
In this test, based upon the structure as illustrated in Fig~\ref{fig:RFmodel}, four features including AMMC, StoLR, viscosity and CG are utilised as the inputs of RF model, while the labelled electrode mass load is used as model's output. Then the detailed results of FI, correlations, RF-based model classification and performance comparison would be given and analysed.   

	\begin{figure}[h]
		\centering
		\includegraphics[angle=0,
		width=0.49\textwidth]{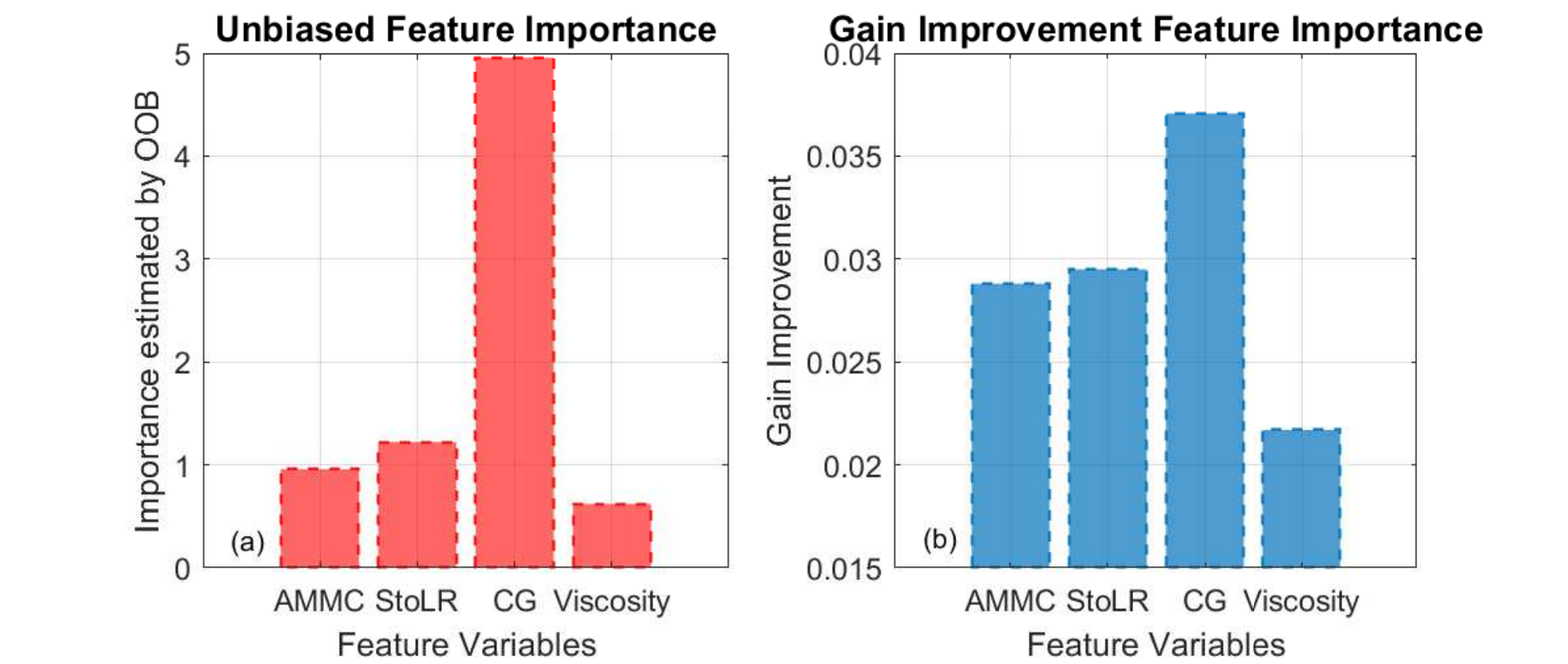}  
		\caption{FI for battery mass load. (a) unbiased FI based on OOB (b) FI based on gain improvement.}
		\label{fig:FIformass} 
	\end{figure}

\subsubsection{Feature analyses} For the mass load classification, following the steps in Workflows~\ref{alg:OOBimportance} and~\ref{alg:Giniimportance}, the quantified unbiased FI as well as gain improvement FI of all four feature variables can be obtained and illustrated in Fig.~\ref{fig:FIformass}. It can be noted that although the value levels between unbiased FI and gain improvement FI are significantly different, they still present the similar trend for all features. Obviously, CG achieves much higher importance values for both unbiased FI (here is 4.78) and gain improvement FI (here is 0.037), indicating that this variable is the most important feature for mass load classification. StoLR and AMMC provide the second and third larger values of both unbiased FI (here are 1.18 and 0.91 respectively) as well as gain improvement FI (here are 0.029 and 0.028 respectively). The viscosity variable presents the smallest values with 0.67 unbiased FI and 0.022 gain improvement FI, indicating that this feature contributes the least to the classification of electrode mass load.

	\begin{figure}[h]
		\centering
		\includegraphics[angle=0,
		width=0.36\textwidth]{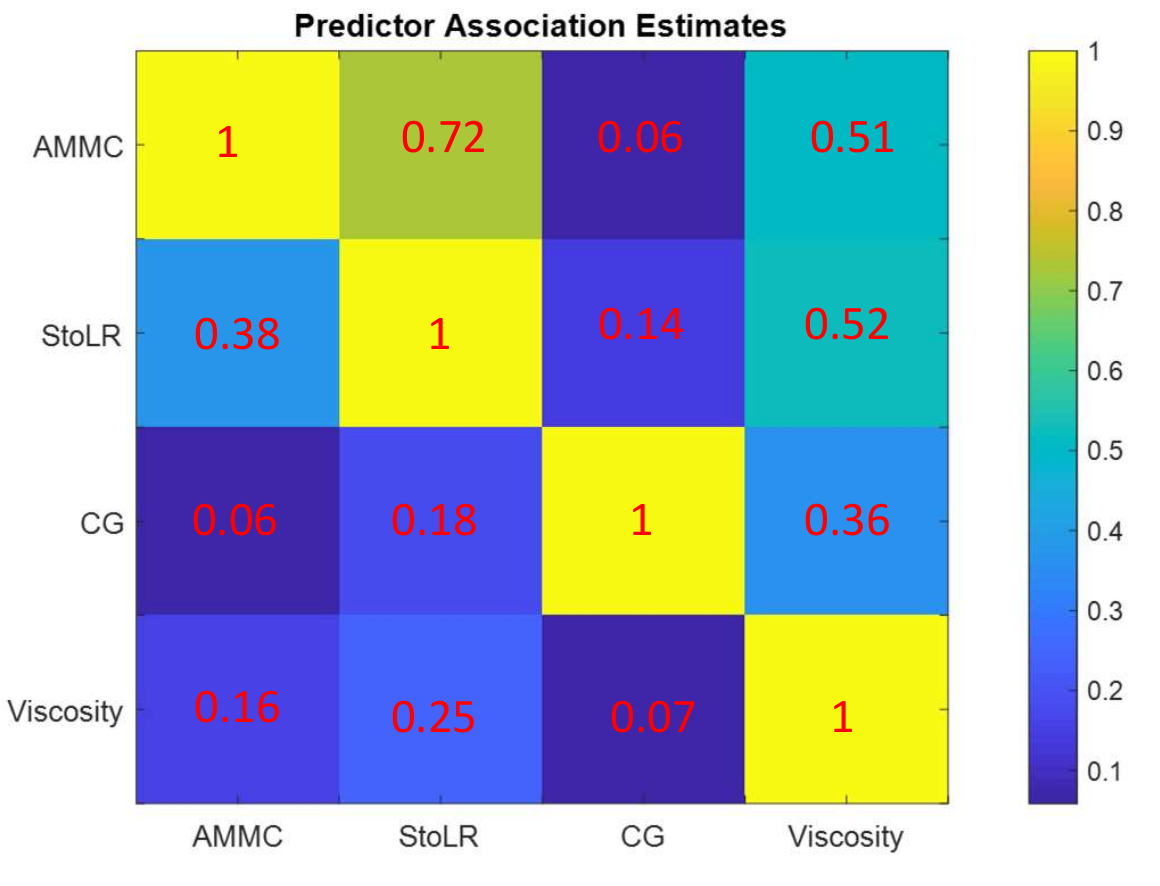} 
		\caption{Heat map to reflect feature correlations for battery mass load case.}
		\label{fig:Fassoformass} 
	\end{figure}	
	
The PMOAs of all feature pairs are calculated next to evaluate the correlations among four features for mass load case. From the heat map in	Fig.~\ref{fig:Fassoformass}, the largest correlation occurs between AMMC and StoLR with a PMOA of 0.72. This correlation output is very useful as the obtained result is consistent with the conclusion from experimental works \cite{kendrick2019advancements}, but we demonstrate how a RF machine-learning framework can aid the interpretation of correlations among feature variables of interest, which could give engineers a guidance to efficiently understand their battery manufacturing chain.


\subsubsection{RF-based model}
\begin{figure}[h]
		\centering
		\includegraphics[angle=0,
		width=0.38\textwidth]{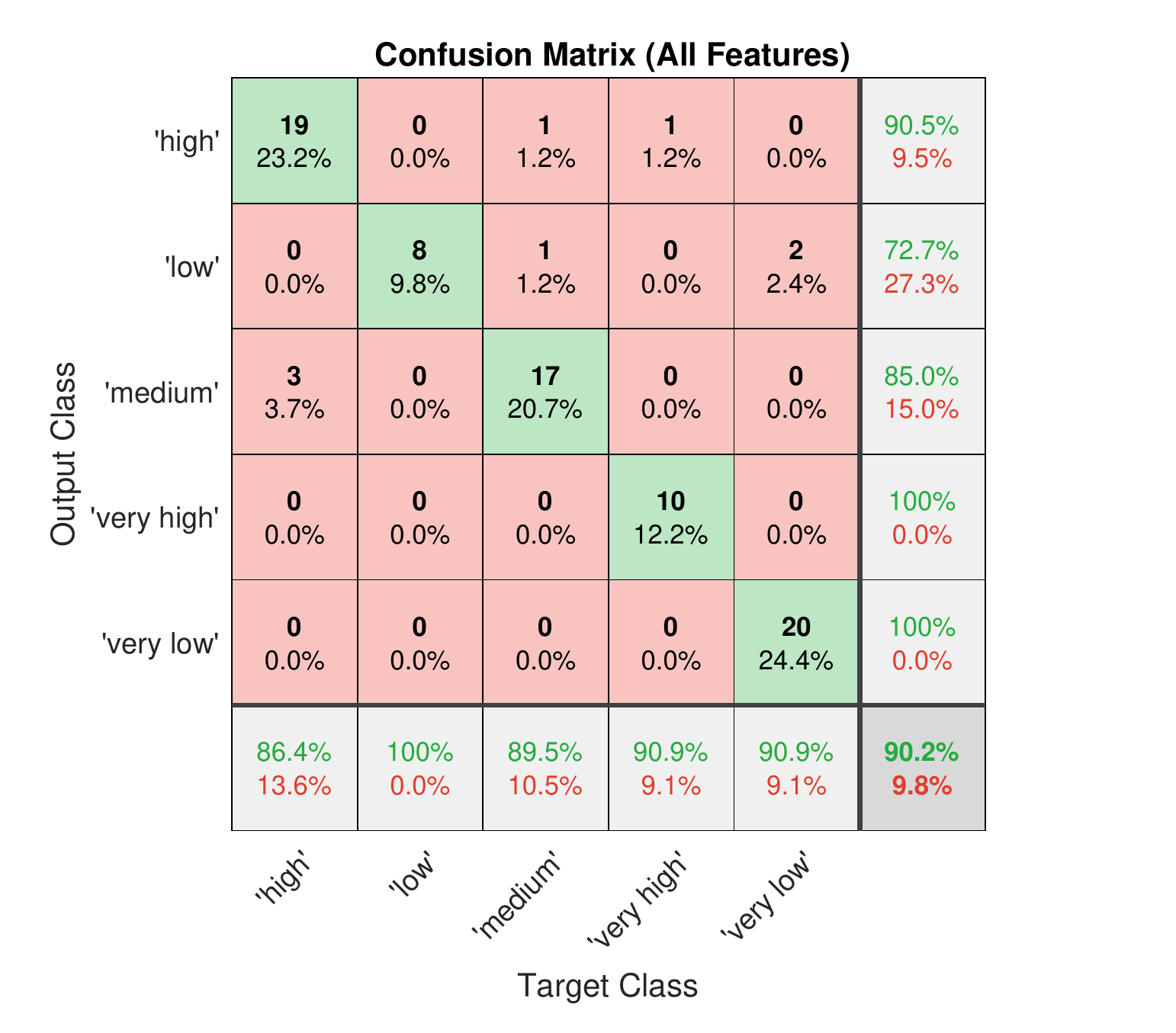}  
		\caption{Confusion matrix for mass load results by using all features.}
		\label{fig:BestMass} 
	\end{figure}

To evaluate the mass load classification results of our proposed RF framework, prediction test through using all features is first carried out. According to the corresponding CM in Fig.~\ref{fig:BestMass}, a satisfactory $OCCrate$ with 90.2\%  is achieved. Quantitatively, the classes 'very high' and 'very low' achieve 100\% $Prate$. The worst classification result is the 'low' class with 72.7\% $Prate$. This is mainly caused by 2 observations are incorrectly classified as 'very low' and 1 observation is classified as 'medium'.

	\subsubsection{Performance comparison}
	\begin{figure}[h]
		\centering
		\includegraphics[angle=0,
		width=0.5\textwidth]{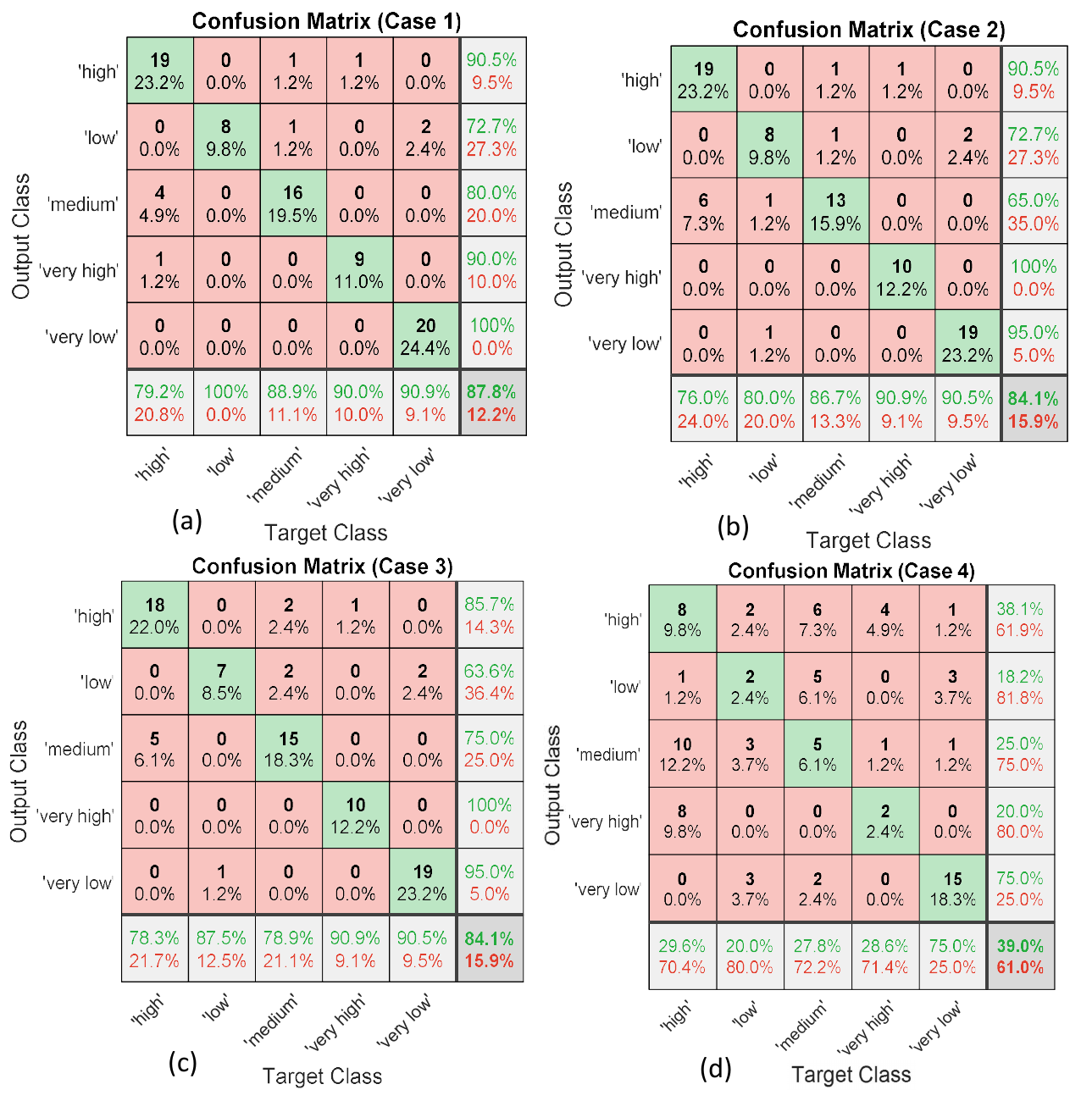}  
		\caption{Confusion matrices for mass load classification results for different cases: (a) Case 1, (b) Case 2, (c) Case 3, (d) Case 4.}
		\label{fig:CMforporoMassl} 
	\end{figure}
	
	\begin{table}[h]
		\caption{Quantitative performance metrics for battery electrode mass load classification} 
		\centering
		\begin{tabular}{p{3.0cm}<\centering  p{1cm}<\centering  p{1cm}<\centering  p{1.5cm}<\centering  
				} 
			\hline\hline   
			\rule{-3pt}{1.0\normalbaselineskip}
			Cases.      & $macroP$ & $macroR$ & $macroF1$ \\ 
			\hline		    
			All features	 & 89.6\%	 & 91.5\%	& 90.1\%  \\
			Case 1  &86.6\%  & 89.8\% & 90.0\%  \\
			Case 2	 & 84.6\%	 & 	84.8\%		& 84.6\%  \\	
			Case 3 	 & 	83.9\%  & 85.2\%	 & 	84.3\% 	\\			   
			Case 4	 & 35.3\% & 36.2\% & 35.4\%\\
			\hline\hline 	
		\end{tabular}
		\label{table:Massloadresults}
	\end{table}	
	
Next, to further investigate the effects of each feature on the mass load classification results, four different cases with various combinations of three features are tested and compared. Specifically, Case 1 consists of CG, AMMC and StoLR features. Case 2 contains CG, AMMC and viscosity features. Case 3 includes CG, StoLR and viscosity features. Case 4 is composed of AMMC, StoLR and viscosity features. Fig.~\ref{fig:CMforporoMassl} and Table~\ref{table:Massloadresults} illustrate the corresponding CMs and performance metrics of all cases. It can be seen that Case 1 provides the best classification results with 86.6\% $macroP$, 89.8\% $macroR$ and 90.0\% $macroF1$, which are only 3.3\%, 1.9\% and 0.1\% less than those from the case of all features. This implies that using CG, AMMC and StoLR is sufficient for mass load classification. Cases 2 and 3 provide the similar performance metrics, which indicates that similar effects exist between AMMC and StoLR. Interestingly, without involving CG, the performance metrics of Case 4 largely decrease, indicating that CG plays a significantly important role in the mass load classification.

\subsection{RF classification model for battery porosity}

Next, the battery electrode porosity classification test is also conducted. The inputs of this test are the same as those from mass load test, while the output here becomes porosity. 

\subsubsection{Feature analyses}

\begin{figure}[h]
		\centering
		\includegraphics[angle=0,
		width=0.46\textwidth]{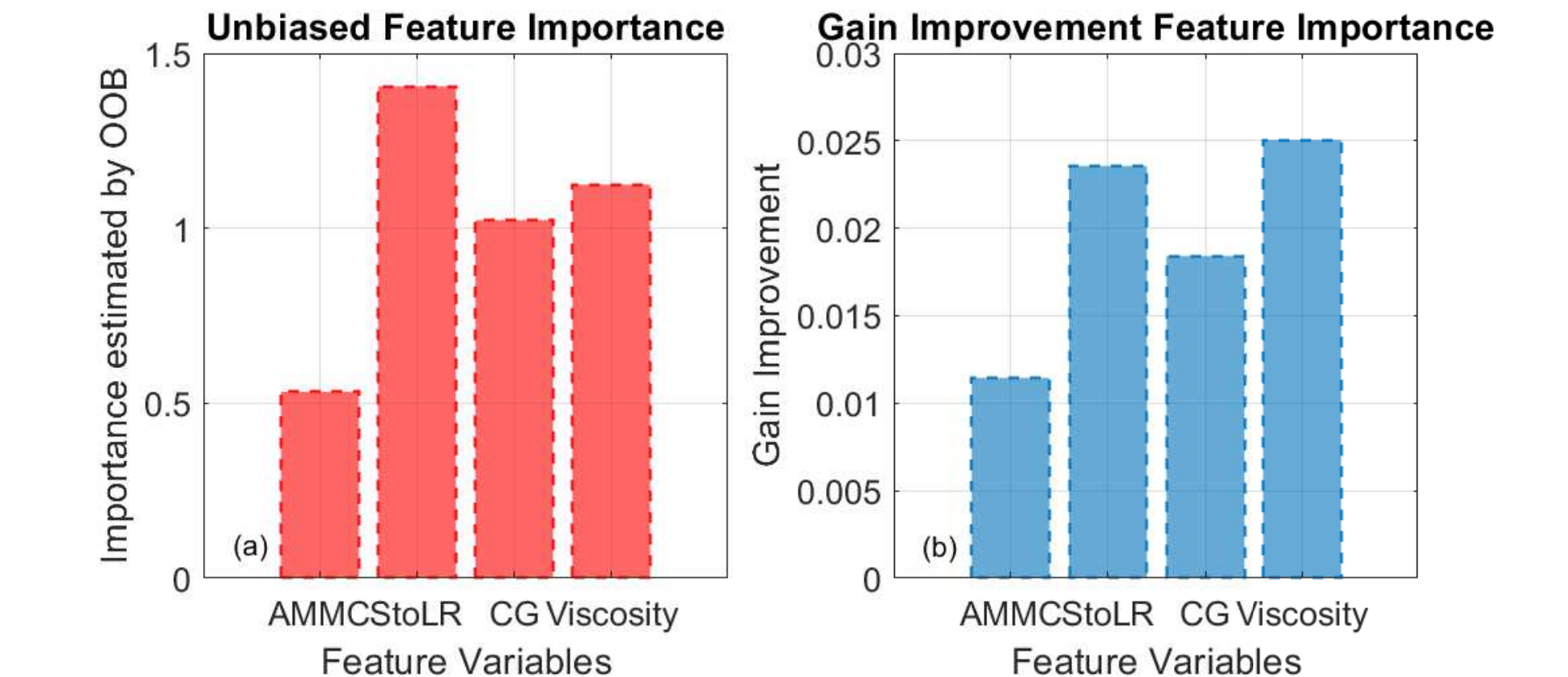}  
		\caption{Feature importance for battery porosity. (a) unbiased FI based on OOB (b) FI based on gain improvement.}
		\label{fig:FIforporo} 
	\end{figure}

\begin{figure}[h]
		\centering
		\includegraphics[angle=0,
		width=0.34\textwidth]{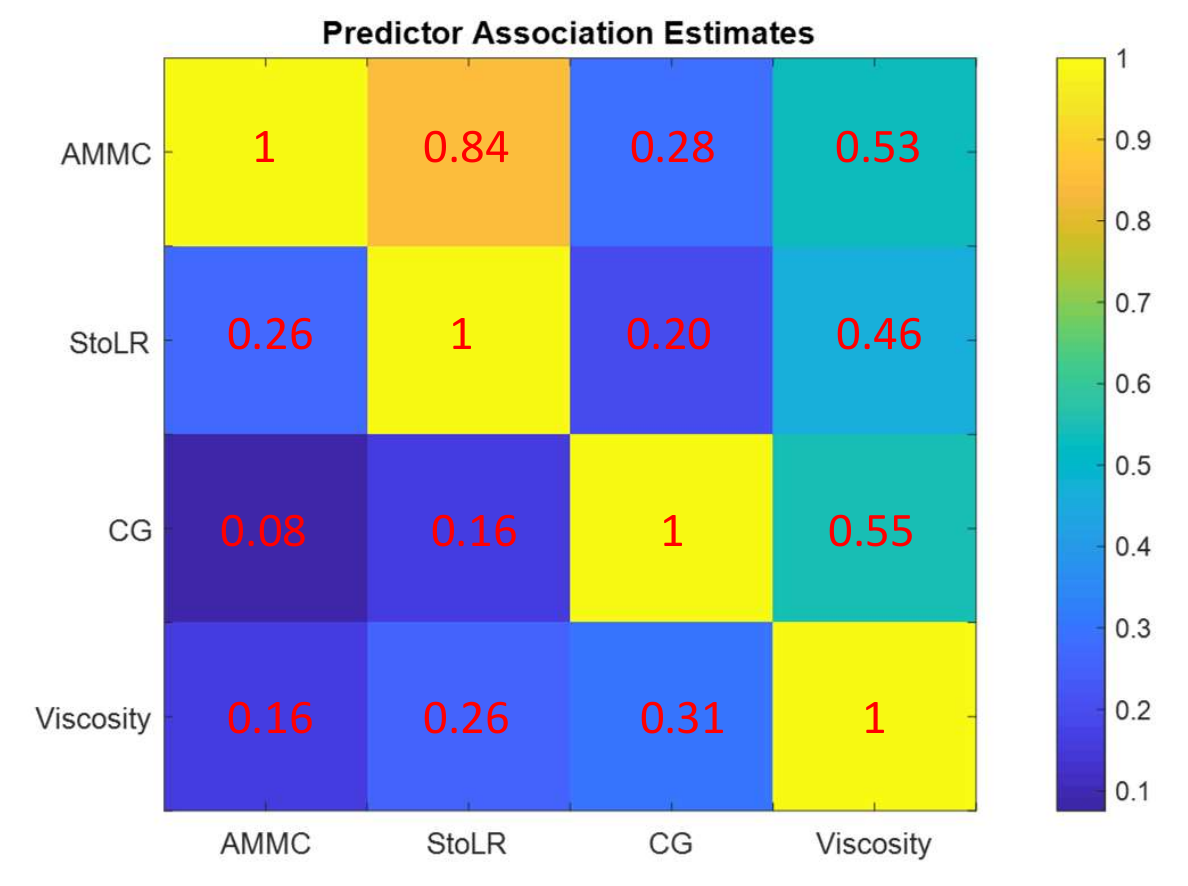}  
		\caption{Heat map to reflect feature correlations for battery porosity case.}
		\label{fig:Fassoforporo} 
	\end{figure}

Fig.~\ref{fig:FIforporo} illustrates the corresponding unbiased FI and gain improvement FI. The metrics indicate that StoLR and viscosity are the two most contributing features while AMMC is the worst one. Next, from the association estimates of corresponding feature pairs in  Fig.~\ref{fig:Fassoforporo}, one PMOA of the AMMC-StoLR pair presents the highest value with 0.84, indicating that these two features may have strong potential correlations for battery electrode porosity classification case.

\subsubsection{RF-based model}

\begin{figure}[h]
		\centering
		\includegraphics[angle=0,
		width=0.38\textwidth]{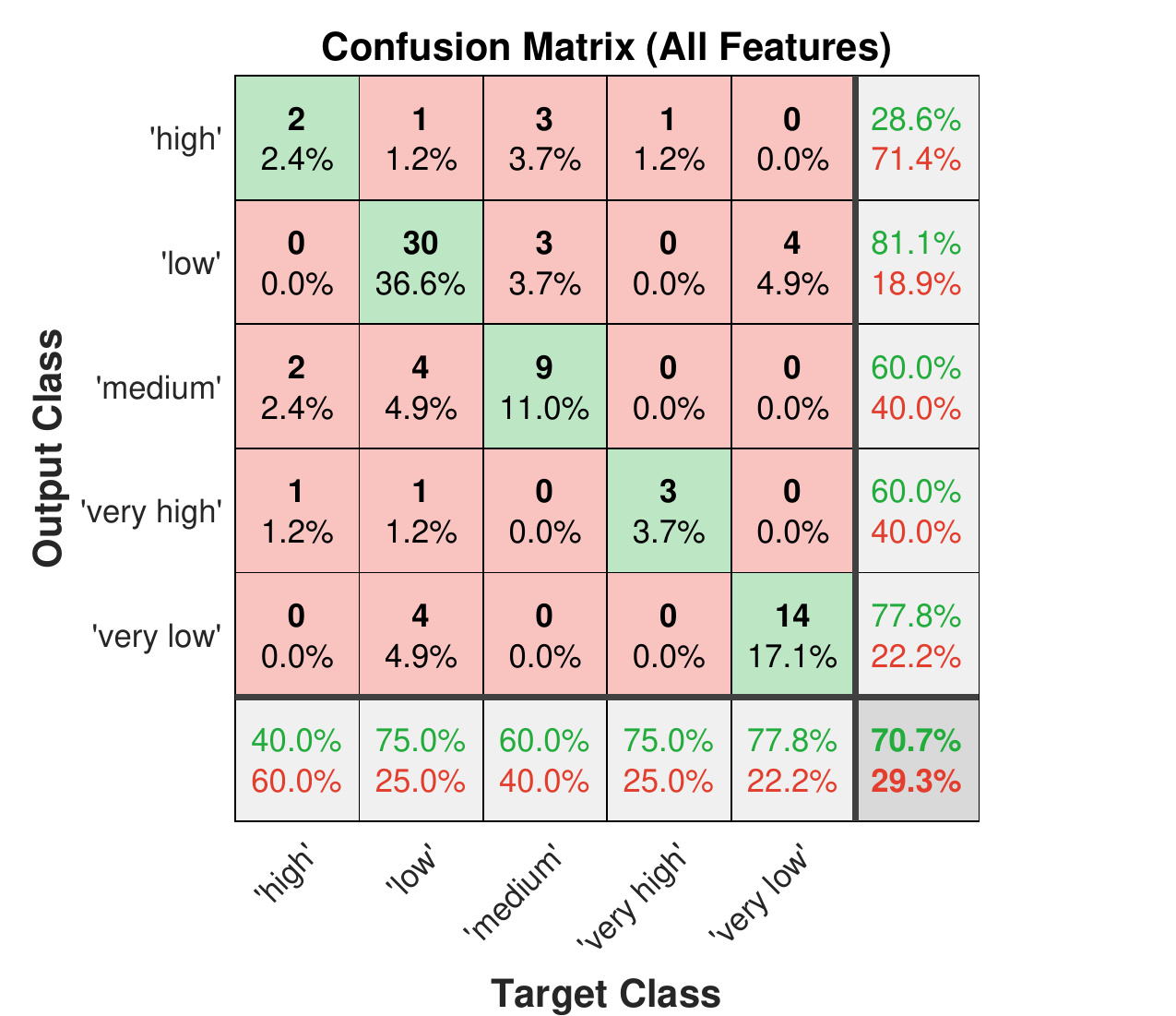}  
		\caption{Confusion matrix for porosity results by using all features.}
		\label{fig:CMforporo} 
	\end{figure}

	Fig.~\ref{fig:CMforporo} illustrates the CM for the porosity classification results when using all features. This test achieves a classification result with 70.7\% $OCCrate$, which is mainly caused by several misclassified results such as those with class label 'high'. In comparison with battery mass load case, it can be concluded that these features cannot fully and well determine the qualities of electrode porosity.

\subsubsection{Performance comparison} 
\begin{figure}[h]
		\centering
		\includegraphics[angle=0,
		width=0.5\textwidth]{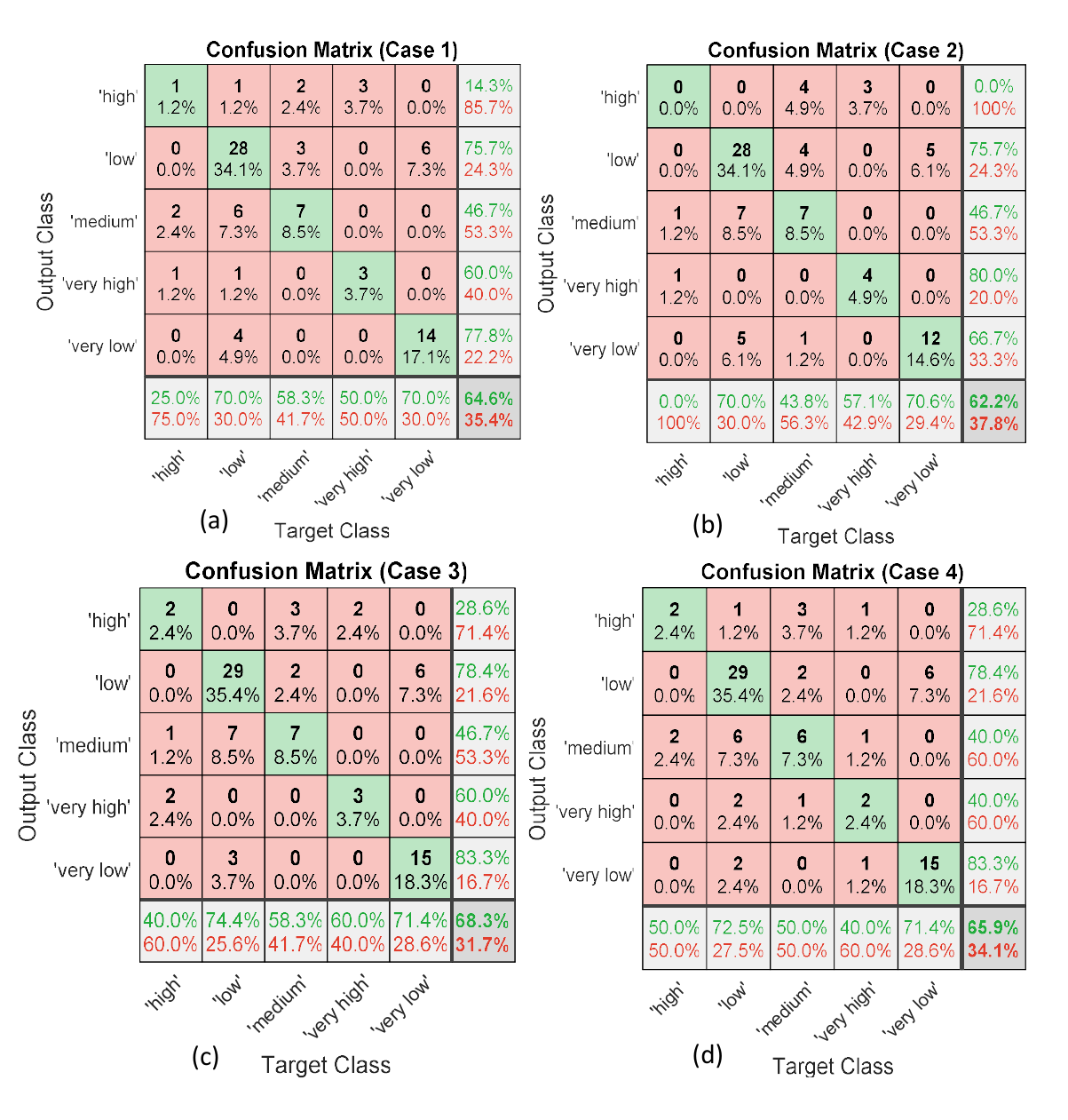}  
		\caption{Confusion matrices for porosity classification results for different cases: (a) Case 1, (b) Case 2, (c) Case 3, (d) Case 4.}
		\label{fig:CMforporoCase} 
	\end{figure}

To further investigate the influence of these features on the electrode porosity classifications, four tests with the same feature combination cases as those from mass load are compared here. Their corresponding CMs and performance metrics are shown in Fig.~\ref{fig:CMforporoCase} and Table~\ref{table:porosityresults}, respectively. Specifically, by using the three most important features (StoLR, CG and viscosity), Case 3 achieves the best classification results with 59.4\% $macroP$, 60.8\% $macroR$, 59.7\% $macroF1$ and 68.3\% $OCCrate$. In contrast, using AMMC to replace any other three features, the related classification performance is reduced accordingly. However, the overall porosity classification results are all worse than those from mass load cases. These facts signify that for battery electrode porosity, more other related IPFs and PPs are recommended to be considered for further improving its classification performance.


\begin{table}[h]
		\caption{Quantitative performance metrics for battery electrode porosity classification} 
		\centering
		\begin{tabular}{p{3.0cm}<\centering  p{1cm}<\centering  p{1cm}<\centering  p{1.5cm}<\centering  
				} 
			\hline\hline   
			\rule{-3pt}{1.0\normalbaselineskip}
			Cases.      & $macroP$ & $macroR$ & $macroF1$ \\ 
			\hline		    
			All features	 & 61.5\%	 & 65.6\%	& 66.4\%  \\
			Case 1  &54.9\%  & 54.7\% & 54.9\%  \\
			Case 2	 & 53.8\%	 & 	48.3\%		& 50.6\%  \\	
			Case 3 	 & 	59.4\%  & 60.8\%	 & 	59.7\% 	\\	   
			Case 4	 & 67.2\% & 56.8\% & 54.6\%\\
			\hline\hline 	
		\end{tabular}
		\label{table:porosityresults}
	\end{table}

\subsection{Discussions}
In this subsection, two tests are designed to investigate the hyper-parameters tuning and compare the performance of proposed RF  with other typical classification methods, followed by the further discussions of results in subsections IV-$A$ and IV-$B$. 

\subsubsection{Hyper-parameters tuning}
As mentioned in subsection~\ref{S3}-C, for the RF classification model, $J$ and $m$ are two key hyper-parameters required to be carefully tuned. Through setting up random combinations to train model and score the mean cross-validated accuracy (MCVA), the randomized search solution \cite{bergstra2012random} is utilised here to determine suitable values of $J$ and $m$ for both electrode mass load and porosity classification cases. Based upon the Python module Scikit-learn with a 2.40 GHz Intel Pentium 4 CPU, the randomized search can be conveniently carried out by using the function module $RandomizedSearchCV$. In our study, the search range of $J$ is set as: $range(40,120,20)$, while the candidates of $m$ is $[2,3]$, respectively. Table~\ref{table:randomsearch} illustrates the classification performance with various hyper-parameter combinations. It can be seen that $J=100,m=3$ presents the best MCVA with 90.2\% for the mass load case, while $J=80,m=3$ provides the best MCVA with 70.6\% for the porosity case. Therefore, the related RF classification models are set with these optimised hyper-parameters in our study.

\begin{table}[h]
		\caption{Results of hyper-parameters tuning} 
		\centering
		\begin{tabular}{p{3.5cm}<\centering  p{1.5cm}<\centering  p{1.5cm}<\centering 
				} 
			\hline\hline   
			\rule{-3pt}{1.0\normalbaselineskip}
			Hyper-parameter combinations      & MCVA (mass load) & MCVA (porosity) \\ 
			\hline		    
			$J=40,m=2$	 & 88.0\%	 & 62.2\%	  \\
			$J=60,m=2$  &88.5\%  & 64.8\%  \\
			$J=80,m=2$	 & 88.8\%	 & 	66.1\%	 \\	
			$J=100,m=2$ & 	89.8\%  & 68.1\% 	\\	   
			$J=120,m=2$	 & 90.0\% & 69.3\% \\
			$J=40,m=3$	 & 88.6\% & 62.5\% \\
			$J=60,m=3$	 & 69.1\% & 65.3\% \\
			$J=80,m=3$	 & 90.0\% & \textbf{70.6\%} \\
			$J=100,m=3$	 & \textbf{90.2\%} & 70.5\% \\
			$J=120,m=3$	 & 90.2\% & 70.6\% \\
			\hline\hline 	
		\end{tabular}
		\label{table:randomsearch}
	\end{table}	
	
\subsubsection{Comparisons with other approaches}	

To further reflect the effectiveness of our designed RF model, another three popular classification approaches including the DT, k-nearest neighbors (KNN) and support vector machine (SVM) are utilised as the benchmarks for comparison purpose. Specifically, DT is a solo CART. KNN belongs to an instance-based learning method and relies on the distance for classification. SVM is a kernel-based method to map inputs into high dimensional spaces for classification~\cite{zhang2016comparison}. Without the loss of generality, randomized search solution is also utilised here to tune their hyper-parameters. After optimisation, DT has the maximum splits number of 20; The neighbors number of KNN is 1; SVM uses the Gaussian kernel with a kernel scale of 0.5. To quantify their classification performance, two significant metrics including the $macroF1$ and the area under curve (AUC) of receiver operating characteristic are utilised. Here the AUC could give the degree or measure of separability of the classes~\cite{biau2016random}. Table~\ref{table:methodcomparison} illustrates the classification results of all these approaches after 5-folds cross-validation. It can be seen that DT shows the worst results, while SVM and RF provide good classification results for both mass load and porosity cases (here RF provides a slightly better $macroF1$ and AUC). Therefore, due to the ensemble learning nature, our proposed RF framework presents competent performance in the classification applications of battery manufacturing.

	\begin{table}[h]
		\caption{Classification results using various approaches} 
		\centering
		\begin{tabular}{p{1.5cm}<\centering  p{1.5cm}<\centering  p{1.1cm}<\centering  p{1.5cm}<\centering p{1.1cm}<\centering  
				} 
			\hline\hline   
			\rule{-3pt}{1.0\normalbaselineskip}
		&	\multicolumn{2}{c}{Mass load} & \multicolumn{2}{c}{Porosity} \\
		\hline
			Approaches      & $macroF1$ & AUC& $macroF1$ & AUC \\ 
			\hline		    
			DT	 & 74.6\% & 0.82 & 53.2\% & 0.77  \\
			KNN  & 83.9\% & 0.92 & 56.4\% & 0.81  \\
			SVM	 & 89.8\% & \textbf{0.98} & 66.0\% & 0.93  \\	
			proposed RF & \textbf{90.1\%} & \textbf{0.98} & \textbf{66.4\%} & \textbf{0.94} 	\\			   
			\hline\hline 	
		\end{tabular}
		\label{table:methodcomparison}
	\end{table}	

\subsubsection{Further discussions}	

Due to the lack of exploiting interpretable data-driven solutions for feature analyses and modelling within the battery manufacturing chain, this paper develops a RF-based framework to quantify variable correlations and importance in the classification of battery electrode properties. According to the obtained results from subsections IV-$A$ and IV-$B$, the electrode mass load can be well determined by the investigated four features (here the $macroF1$ is 90.1\%) while CG plays the most important role in its classification results (nearly 60.7\% decrease). This result is expected as CG would significantly affect the coating weight and thickness, and these coating properties highly determine the electrode mass load. For the results of electrode porosity, the $macroF1$ here is just 66.4\%, indicating that more other feature variables should be considered to better classify the electrode porosity. This result is expected as electrode porosity would be also highly affected by the drying parameters (rate, temperature, pressure, etc) in theory. Not surprisingly, AMMC and StoLR present high correlation for both mass load and porosity cases. This is mainly due to the mass ratio between slurry solid components and slurry mass has strong and direct relations with the active material properties. In contrast, there are not so direct relations for other feature pairs, which leads their correlations become less. Besides, the mass content of active material cannot highly affect the electrode physical property such as porosity, which makes the AMMC here become the less important feature. In light of this, to further improve our proposed RF-based framework for better prediction of electrode porosity, more feature variables from drying and calendering processes such as drying rate, temperature, pressure and calendering speed should be considered. Besides, more available data from other key production processes could be also collected to improve the interpretability of RF model for better understanding battery manufacturing.

\section{Conclusion}
\label{S5}
As battery manufacturing is crucial for determining battery performance, the effective feature analyses and electrode properties classification within manufacturing chain are strongly required. In this article, through using the improved RF technique, a powerful data-driven framework  is designed to not only quantify the importance levels of four key battery manufacturing features but also provide their feature association estimates. The effects of AMMC, StoLR, CG and viscosity on the classifications of both electrode mass load and porosity are all evaluated and analysed. Due to the superiority in terms of interpretability and data-driven nature, the proposed RF classification framework could be conveniently extended to consider more input features from other key manufacturing stages such as mixing, drying, and calendering. As collecting battery manufacturing data requires specific equipment and is time-consuming, our future work would focus on designing extra experiments to generate more related data such as the mixing kneading intensity and speed, the drying rate, temperature and pressure, and the calendering speed, then to further improve the usability of such ML method and accelerate the development of high-performance Li-ion batteries.

\label{section5}
\bibliographystyle{IEEEtranTIE}
\bibliography{IEEEabrv,BIB_1x-TIE-2xxx}\ 

\end{document}